\documentclass{article}
\usepackage{iwpt97}
\usepackage{fullname}
\newcounter{examplectr}
\newcounter{subexamplectr}
\newenvironment{ex}%
   {\vspace{.1in}\addtocounter{examplectr}{1}
     \setcounter{subexamplectr}{0}
     \begin{list}
       {(\arabic{examplectr})}%
       {\setlength{\topsep}{0in}
        \setlength{\leftmargin}{.25in}
               \setlength{\labelsep}{0.075in}}
       \item \begin{minipage}[t]{13.5cm} 
   }%
   {\end{minipage}
    \end{list}\vspace{.1in}}
\newenvironment{subex}%
   { \addtocounter{subexamplectr}{1}
     \begin{list}
       {\alph{subexamplectr}}%
       {\setlength{\topsep}{-\parskip}
        \setlength{\leftmargin}{0.175in}
        \setlength{\labelsep}{0.075in}}
       \item
   }%
   {\end{list}}
\newcommand{\exnum}[2]{\addtocounter{examplectr}{#1}(\arabic{examplectr}{#2})\addtocounter{examplectr}{-#1}}

\title{ENCODING FREQUENCY INFORMATION IN\\
LEXICALIZED GRAMMARS}

\author{John Carroll  \qquad David Weir\\
School of Cognitive and Computing Sciences\\
University of Sussex\\
Falmer, Brighton, BN1 9QH, UK\\
{\normalsize \{johnca,davidw\}@cogs.susx.ac.uk}
}

\begin{document}
\noindent
{\it In Proceedings of the 5th ACL/SIGPARSE International
Workshop on Parsing Technologies, MIT, Cambridge MA, 1997.}
\vspace{-6mm}
\maketitle
\begin{abstract}
We address the issue of how to associate frequency information with
lexicalized grammar formalisms, using Lexicalized Tree Adjoining
Grammar as a representative framework. We consider systematically a number
of alternative probabilistic frameworks, evaluating their adequacy from
both a theoretical and empirical perspective using data from existing
large treebanks. We also propose three orthogonal approaches for backing off
probability estimates to cope with the large number of parameters involved.
\end{abstract}

\section{Introduction}
\label{sec-intro}

When performing a derivation with a grammar it is usually the case that,
at certain points in the derivation process, the grammar licenses
several alternative ways of continuing with the derivation.  In the case
of context-free grammar~(CFG) such nondeterminism arises when there are
several productions for the nonterminal that is being rewritten.
Frequency information associated with the grammar may be used to assign
a probability to each of the alternatives.  In general, it must always
be the case that at every point where a choice is available the
probabilities of all the alternatives sum to $1$. This frequency
information provides a parser with a way of dealing with the problem
of ambiguity: the parser can use the information either to preferentially
explore possibilities that are more likely, or to assign probabilities to
the alternative parses.

There can be many ways of associating frequency information with the
components making up a grammar formalism.  For example, just two of the
options in the case of CFG are: (1)~associating a single probability with
each production that determines the probability of its use wherever it
is applicable (i.e.\ Stochastic CFG; SCFG~\cite{booth73}); or
(2)~associating different probabilities with a production depending on the
particular nonterminal occurrence (on the RHS of a production) that is being
rewritten~\cite{chitrao90}.  In the latter case probabilities depend on the
context (within a production) of the nonterminal being rewritten. In
general, while there may be alternative ways of associating frequency
information with grammars, the aim is always to provide a way of
associating probabilities with alternatives that arise during derivations.

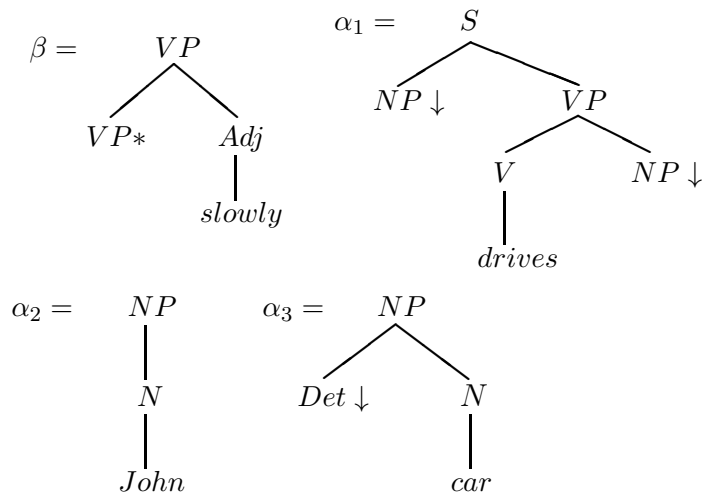
\begin{figure}[bt]
\begin{center}
\setlength{\unitlength}{0.00062475in}%
\begingroup\makeatletter\ifx\SetFigFont\undefined
\def\x#1#2#3#4#5#6#7\relax{\def\x{#1#2#3#4#5#6}}%
\expandafter\x\fmtname xxxxxx\relax \def\y{splain}%
\ifx\x\y   
\gdef\SetFigFont#1#2#3{%
  \ifnum #1<17\tiny\else \ifnum #1<20\small\else
  \ifnum #1<24\normalsize\else \ifnum #1<29\large\else
  \ifnum #1<34\Large\else \ifnum #1<41\LARGE\else
     \huge\fi\fi\fi\fi\fi\fi
  \csname #3\endcsname}%
\else
\gdef\SetFigFont#1#2#3{\begingroup
  \count@#1\relax \ifnum 25<\count@\count@25\fi
  \def\x{\endgroup\@setsize\SetFigFont{#2pt}}%
  \expandafter\x
    \csname \romannumeral\the\count@ pt\expandafter\endcsname
    \csname @\romannumeral\the\count@ pt\endcsname
  \csname #3\endcsname}%
\fi
\fi\endgroup
\begin{picture}(5352,4071)(526,-3538)
\thicklines
\put(3742,-2074){\line(-4,-3){600}}
\put(3751,-2086){\line( 4,-3){600}}
\put(4368,-2836){\line( 0,-1){450}}
\put(1651,-2086){\line( 0,-1){450}}
\put(1651,-2836){\line( 0,-1){450}}
\put(2401,-661){\line( 0,-1){375}}
\put(1886,101){\line( 6,-5){531.148}}
\put(1886,101){\line(-6,-5){531.148}}
\put(4368,287){\line(-5,-3){606.618}}
\put(4368,287){\line( 5,-2){905.172}}
\put(5266,-331){\line(-2,-1){600}}
\put(5266,-331){\line( 2,-1){600}}
\put(4651,-961){\line( 0,-1){450}}
\put(3226,389){\makebox(0,0)[lb]{\smash{\SetFigFont{11}{14.4}{rm}$\alpha_1=$}}}
\put(4276,389){\makebox(0,0)[lb]{\smash{\SetFigFont{11}{14.4}{rm}$S$}}}
\put(3586,-2011){\makebox(0,0)[lb]{\smash{\SetFigFont{11}{14.4}{rm}$NP$}}}
\put(2921,-2761){\makebox(0,0)[lb]{\smash{\SetFigFont{11}{14.4}{rm}$Det\downarrow$}}}
\put(4276,-2761){\makebox(0,0)[lb]{\smash{\SetFigFont{12}{14.4}{rm}$N$}}}
\put(4201,-3471){\makebox(0,0)[lb]{\smash{\SetFigFont{11}{14.4}{rm}$car$}}}
\put(2626,-2011){\makebox(0,0)[lb]{\smash{\SetFigFont{11}{14.4}{rm}$\alpha_3=$}}}
\put(526,-2011){\makebox(0,0)[lb]{\smash{\SetFigFont{11}{14.4}{rm}$\alpha_2=$}}}
\put(1501,-2011){\makebox(0,0)[lb]{\smash{\SetFigFont{11}{14.4}{rm}$NP$}}}
\put(1576,-2761){\makebox(0,0)[lb]{\smash{\SetFigFont{11}{14.4}{rm}$N$}}}
\put(1421,-3471){\makebox(0,0)[lb]{\smash{\SetFigFont{11}{14.4}{rm}$John$}}}
\put(2101,-1221){\makebox(0,0)[lb]{\smash{\SetFigFont{11}{14.4}{rm}$slowly$}}}
\put(2251,-586){\makebox(0,0)[lb]{\smash{\SetFigFont{11}{14.4}{rm}$Adj$}}}
\put(1151,-586){\makebox(0,0)[lb]{\smash{\SetFigFont{11}{14.4}{rm}$VP*$}}}
\put(1726,164){\makebox(0,0)[lb]{\smash{\SetFigFont{11}{14.4}{rm}$VP$}}}
\put(676,164){\makebox(0,0)[lb]{\smash{\SetFigFont{11}{14.4}{rm}$\beta=$}}}
\put(5126,-286){\makebox(0,0)[lb]{\smash{\SetFigFont{11}{14.4}{rm}$VP$}}}
\put(5706,-886){\makebox(0,0)[lb]{\smash{\SetFigFont{11}{14.4}{rm}$NP\downarrow$}}}
\put(4556,-886){\makebox(0,0)[lb]{\smash{\SetFigFont{11}{14.4}{rm}$V$}}}
\put(4426,-1596){\makebox(0,0)[lb]{\smash{\SetFigFont{11}{14.4}{rm}$drives$}}}
\put(3551,-286){\makebox(0,0)[lb]{\smash{\SetFigFont{11}{14.4}{rm}$NP\downarrow$}}}
\end{picture}

\end{center}
\caption{An Example Grammar}
\label{fig-ltag}
\end{figure}

\begin{figure}[bt]
\begin{center}
\setlength{\unitlength}{0.00072475in}%
\begingroup\makeatletter\ifx\SetFigFont\undefined
\def\x#1#2#3#4#5#6#7\relax{\def\x{#1#2#3#4#5#6}}%
\expandafter\x\fmtname xxxxxx\relax \def\y{splain}%
\ifx\x\y   
\gdef\SetFigFont#1#2#3{%
  \ifnum #1<17\tiny\else \ifnum #1<20\small\else
  \ifnum #1<24\normalsize\else \ifnum #1<29\large\else
  \ifnum #1<34\Large\else \ifnum #1<41\LARGE\else
     \huge\fi\fi\fi\fi\fi\fi
  \csname #3\endcsname}%
\else
\gdef\SetFigFont#1#2#3{\begingroup
  \count@#1\relax \ifnum 25<\count@\count@25\fi
  \def\x{\endgroup\@setsize\SetFigFont{#2pt}}%
  \expandafter\x
    \csname \romannumeral\the\count@ pt\expandafter\endcsname
    \csname @\romannumeral\the\count@ pt\endcsname
  \csname #3\endcsname}%
\fi
\fi\endgroup
\begin{picture}(4149,2349)(664,-1798)
\thicklines
\put(4050,-306){\circle{25}}
\put(4050,-306){\circle{15}}
\put(3176,-586){\circle{25}}
\put(4501,-586){\circle{25}}
\put(3595,606){\line(-1,-1){1194.500}}
\put(2401,-586){\line( 1, 0){2400}}
\put(4801,-586){\line(-1, 1){1198.500}}
\put(1276,-361){\line(-1, 0){600}}
\put(676,-361){\line( 4, 5){600}}
\put(1801,-961){\line( 5,-6){565.869}}
\put(2375,-1636){\line(-1, 0){1405}}
\put(976,-1636){\line( 5, 6){577.869}}
\put(4201,-1111){\line( 5,-6){565.869}}
\put(4775,-1786){\line(-1, 0){1405}}
\put(3376,-1786){\line( 5, 6){577.869}}
\put(1501,-361){\line( 1, 0){675}}
\put(2176,-361){\line(-5, 6){645.492}}
\put(1786,250){\vector( 4,-1){2169.412}}
\put(1861,-936){\vector( 4, 1){1196.500}}
\put(4091,-961){\vector( 1, 1){322.500}}
\put(1271,329){\makebox(0,0)[lb]{\smash{\SetFigFont{11}{14.4}{rm}$VP$}}}
\put(1271,-426){\makebox(0,0)[lb]{\smash{\SetFigFont{11}{14.4}{rm}$VP$}}}
\put(1541,-1021){\makebox(0,0)[lb]{\smash{\SetFigFont{11}{14.4}{rm}$NP$}}}
\put(3951,-1151){\makebox(0,0)[lb]{\smash{\SetFigFont{11}{14.4}{rm}$NP$}}}
\put(1331,-64){\makebox(0,0)[lb]{\smash{\SetFigFont{11}{14.4}{rm}$\beta$}}}
\put(1591,-1411){\makebox(0,0)[lb]{\smash{\SetFigFont{11}{14.4}{rm}$\alpha_2$}}}
\put(4001,-1531){\makebox(0,0)[lb]{\smash{\SetFigFont{11}{14.4}{rm}$\alpha_3$}}}
\put(3556,-61){\makebox(0,0)[lb]{\smash{\SetFigFont{11}{14.4}{rm}$\alpha_1$}}}
\put(2251,194){\makebox(0,0)[lb]{\smash{\SetFigFont{11}{14.4}{rm}$i_1$}}}
\put(2236,-991){\makebox(0,0)[lb]{\smash{\SetFigFont{11}{14.4}{rm}$i_2$}}}
\put(3991,-831){\makebox(0,0)[lb]{\smash{\SetFigFont{11}{14.4}{rm}$i_3$}}}
\end{picture}

\end{center}
\caption{A Derivation Step}
\label{fig-ex-derived}
\end{figure}
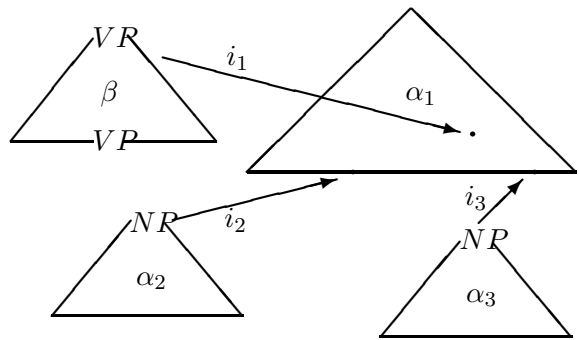

This paper is concerned with how the kind of frequency information that
would be useful to a parser can be associated with lexicalized grammar
formalisms. To properly ground the discussion we will use Lexicalized
Tree Adjoining Grammar (LTAG) as a representative framework, although
our remarks can be applied to lexicalized grammar formalisms more
generally.  We begin by considering the derivation process, and, in
particular, the nature of derivation steps. At the heart of a TAG is a
finite set of trees (the elementary trees of the grammar). In an
LTAG these trees are `anchored' with lexical items and the
tree gives a possible context for its anchor by providing a structure
into which its complements and modifiers can be attached. For example,
Figure~\ref{fig-ltag} shows four elementary trees---one {\it auxiliary}
tree $\beta$ and three {\it initial} trees $\alpha_1$, $\alpha_2$ and
$\alpha_3$.  Nodes marked with asterisks and downarrows are foot and
substitution nodes, respectively. In a derivation these trees are
combined using the operations of substitution and adjunction to produce
a derived tree for a complete sentence.  Figure~\ref{fig-ex-derived}
shows a single derivation step in which $\alpha_2$ and $\alpha_3$ are
substituted at frontier nodes (with addresses $1$ and $2\cdot 2$,
respectively) of $\alpha_1$ and $\beta$ is adjoined at an internal node
of $\alpha_1$ (with address $2$)\protect\footnote{The root of a tree has
the address $\epsilon$. The $i$th daughter (where siblings are ordered from
left to right) of a node with address $a$ has address $a\cdot i$}.

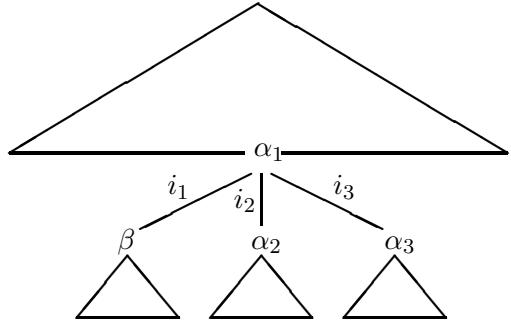
\begin{figure}[bt]
\begin{center}
\setlength{\unitlength}{0.00072475in}%
\begingroup\makeatletter\ifx\SetFigFont\undefined
\def\x#1#2#3#4#5#6#7\relax{\def\x{#1#2#3#4#5#6}}%
\expandafter\x\fmtname xxxxxx\relax \def\y{splain}%
\ifx\x\y   
\gdef\SetFigFont#1#2#3{%
  \ifnum #1<17\tiny\else \ifnum #1<20\small\else
  \ifnum #1<24\normalsize\else \ifnum #1<29\large\else
  \ifnum #1<34\Large\else \ifnum #1<41\LARGE\else
     \huge\fi\fi\fi\fi\fi\fi
  \csname #3\endcsname}%
\else
\gdef\SetFigFont#1#2#3{\begingroup
  \count@#1\relax \ifnum 25<\count@\count@25\fi
  \def\x{\endgroup\@setsize\SetFigFont{#2pt}}%
  \expandafter\x
    \csname \romannumeral\the\count@ pt\expandafter\endcsname
    \csname @\romannumeral\the\count@ pt\endcsname
  \csname #3\endcsname}%
\fi
\fi\endgroup
\begin{picture}(3624,2349)(1789,-1573)
\thicklines
\put(3581,734){\line(-5,-3){1819.853}}
\put(1761,-361){\line( 1, 0){1720}}
\put(3451,-361){\line( 0, 1){  0}}
\put(3451,-361){\line( 0, 1){  0}}
\put(3751,-361){\line( 1, 0){1650}}
\put(5401,-361){\line(-5, 3){1819.853}}
\put(3601,764){\line( 0, 1){  0}}
\put(3601,-511){\line( 0,-1){375}}
\put(3676,-511){\line( 2,-1){810}}
\put(3526,-511){\line(-2,-1){810}}
\put(2626,-1111){\line(-5,-6){375}}
\put(2251,-1561){\line( 1, 0){750}}
\put(3001,-1561){\line(-5, 6){375}}
\put(3601,-1111){\line(-5,-6){375}}
\put(3226,-1561){\line( 1, 0){750}}
\put(3976,-1561){\line(-5, 6){375}}
\put(4576,-1111){\line(-5,-6){375}}
\put(4201,-1561){\line( 1, 0){750}}
\put(4951,-1561){\line(-5, 6){375}}
\put(3546,-381){\makebox(0,0)[lb]{\smash{\SetFigFont{11}{14.4}{rm}$\alpha_1$}}}
\put(2551,-1061){\makebox(0,0)[lb]{\smash{\SetFigFont{11}{14.4}{rm}$\beta$}}}
\put(3526,-1061){\makebox(0,0)[lb]{\smash{\SetFigFont{11}{14.4}{rm}$\alpha_2$}}}
\put(4501,-1061){\makebox(0,0)[lb]{\smash{\SetFigFont{11}{14.4}{rm}$\alpha_3$}}}
\put(2916,-661){\makebox(0,0)[lb]{\smash{\SetFigFont{11}{14.4}{rm}$i_1$}}}
\put(3401,-761){\makebox(0,0)[lb]{\smash{\SetFigFont{11}{14.4}{rm}$i_2$}}}
\put(4126,-661){\makebox(0,0)[lb]{\smash{\SetFigFont{11}{14.4}{rm}$i_3$}}}
\end{picture}

\end{center}
\caption{A Derivation Tree}
\label{fig-ex-derivation}
\end{figure}

When formalizing LTAG derivations, a distinction must be made between
the (object-level) trees that are derived in a derivation and the
(meta-level) trees that are used to fully encode what happens in
derivations.  These trees are referred to as derived and derivation
trees, respectively.  A scheme for encoding TAG derivations was
proposed by \namecite{v87} and later modified by
\namecite{ss94}. Derivation trees show, in
a very direct way, how the elementary trees are combined in
derivations.  Nodes of the derivation trees are labeled by the names
of elementary trees, and edge labels identify tree addresses (i.e.\ node
locations) in elementary trees. Figure~\ref{fig-ex-derivation} shows the
derivation tree resulting from the derivation step in
Figure~\ref{fig-ex-derived}.
The nodes
identified in the derivation tree encode that when the elementary tree
$\alpha_1$ was used, the elementary trees $\beta,\alpha_2,\alpha_3$
were chosen to fit into the various complement and modifier positions.
These positions are identified by the tree addresses
$i_1,i_2,i_3$ labeling the respective edges, where in
this example $i_1=2$, $i_2=1$ and
$i_3=2\cdot 2$\protect\footnote{As \namecite{ss94}
point out matters are somewhat more complex that this.  What we describe
here more closely follows the approach taken by \namecite{rvw95a} in
connection with D-Tree Grammar.}.  In other words, this derivation tree
indicates which choice was made as to how the node $\alpha_1$ should be {\em
expanded}.  In general, there may have been many alternatives since
modification is usually optional and different complements can be selected.

By identifying the nature of nondeterminism in LTAG derivations we have
determined the role that frequency information plays. For each
elementary tree of the grammar, frequency information must somehow
determine how the probability mass is to be distributed among all the
alternative ways of expanding that tree. In section~\ref{sec-approaches} we 
consider a number of ways in which this frequency information can be
associated with a grammar. We then go on to evaluate the degree to which
each scheme can, in principle, distinguish the probability of certain kinds
of derivational phenomena, using data from existing large treebanks
(section~\ref{sec-evaluation}). We discuss in section~\ref{sec-estimation}
how to estimate the large number of probabilistic parameters involved, and
propose three orthogonal approaches for smoothing the probability estimates
obtained. The paper concludes (section~\ref{sec-other}) with comparisons with
other related work.

\section{Frequency Information in Lexicalized
Grammars}\label{sec-approaches}

In this section we consider four ways of associating frequency
information with lexicalized grammars. Using the LTAG framework outlined
in section~\ref{sec-intro} as a basis we define four Stochastic Lexicalized
Grammar formalisms which we will refer to as SLG(1), SLG(2), SLG(3) and
SLG(4). The differences between them lie in how fine-grained the
frequency information is, which in turn determines the extent to which the
resulting probabilities can be dependent on derivational context.

\subsection{Context-Free Frequencies}

The first approach we consider is the simplest and will be referred to
as {\bf SLG(1)}.  A single probability is associated with each
elementary tree.  This is the probability that that tree is used in a
derivation in preference to another tree with the same nonterminal at
its root.  A grammar is therefore well-formed if, for each nonterminal
symbol that can be at the root of a substitutable (adjoinable) tree,
the sum of probabilities associated with all substitutable
(adjoinable) trees with the same root nonterminal is $1$.  When
nondeterminism arises in a derivation nothing about the derivational
context can influence the way that a tree is expanded, since the
probability that the various possible trees are adjoined or substituted
at each node depends only on the identity of the nonterminal at that
node.  As a result we say the frequency information in an SLG(1) is
{\em context-free}.

\subsection{Node-Dependent Frequencies}

The second approach considered here, which we will call {\bf SLG(2)},
has been described before by both \namecite{schabes:1992} and
\namecite{resnik:1992}.  We describe the scheme of Schabes
here, though the approach taken by Resnik is equivalent. In defining
his scheme Schabes uses a stochastic version of a context-free-like
grammar formalism called Linear Indexed Grammar~(LIG).  Based on the
construction used to show the weak equivalence of TAG and
LIG~\cite{vw94}, a LIG is constructed from a given LTAG such that
derivation trees of the LIG encode the derived trees of the associated
LTAG. Compiling LTAG to LIG involves decomposing the elementary trees
into single-level trees and introducing additional productions
explicitly encoding every possible adjunction and substitution
possibility\protect\footnote{This scheme has proved useful in the
  study of LTAG parsing~\cite{schabes:1990,vw93b,boullier96} since
  this pre-compilation process alleviates the need to do what amounts
  to the same decomposition process during parsing.}.  It is the LIG
productions encoding adjunction and substitution possibilities that
are assigned probabilities\protect\footnote{The other productions
  (that decompose the tree structure) are assigned a probability of
  $1$ since they are deterministic.}.  The probabilities associated
with all the productions that encode possible adjunctions
(substitutions) at a node must sum to $1$.  The key feature of these
probability-bearing LIG productions, in the context of the current
discussion, is that they encode the adjunction or substitution of a
specific elementary tree at a specific place in another elementary
tree.  This means that the frequency information can to some extent be
dependent on context.  In particular, when faced with nondeterminism
in the way that some elementary tree is expanded during a derivation,
the probability distribution associated with the alternative
adjunctions or substitutions at a given node can depend on which
elementary tree that node comes from.  As a result we call the
frequency information in SLG(2) {\bf node-dependent}.  This makes
SLG(2) more expressive than SLG(1).  As both Schabes and Resnik point
out, by leveraging LTAG's extended domain of locality this approach
allows probabilities to model both lexical and structural
co-occurrence preferences.

The head automata of \namecite{alshawi:96}
also fit into the SLG(2) formalism since they involve a dependency parameter
which gives the probability that a head has a given word as a
particular dependent.

\subsection{Locally-Dependent Frequencies}

The third approach is {\bf SLG(3)} which falls out quite naturally from
consideration of the TAG derivation process.  As we discussed in
the introduction, LTAG derivations can be encoded with derivation
trees in which nodes are labeled by the names of elementary trees and
edges labeled by the addresses of substitution and adjunction nodes.
The tree addresses can be omitted from derivation trees if a fixed
linear order is established on {\em all} of the adjunction and
substitution nodes in each elementary tree and this ordering is used
to order siblings in the derivation tree.  Given this possibility,
\namecite{vwj87b} have shown that the
set of derivation trees associated with a TAG forms a local set and
can therefore be generated by a context-free grammar
(CFG)\protect\footnote{In such context-free grammars, the terminal and
  nonterminal alphabets are not necessarily disjoint, and only the
  trees generated by the grammar (not their frontier strings) are of
  any interest.}. The productions of this {\bf meta-grammar} encode
possible derivation steps of the grammar.  In other words, each
meta-production encodes one way of (fully) expanding an elementary
tree\protect\footnote{In the formulation of TAG derivations given by
  \namecite{ss94} an arbitrary number of
  modifications can take place at a single node. This means that there
  are an infinite number of productions in the meta-grammar, i.e., an
  infinite number of ways of expanding trees. This means that a pure
  version of SLG(3) is not possible. See Section~\ref{sec-smoothing}
  for ways to deal with this issue.}.  In SLG(3) a probability is
associated with each of these meta-productions.  A SLG(3) is
well-formed if for each elementary tree the sum of the probabilities
associated with the meta-productions for that tree is $1$.

In contrast to SLG(2)---which is limited to giving the probability
that a tree anchored with a given lexical item is substituted or
adjoined into a tree anchored with a second lexical item---SLG(3)
specifies the probability that a particular {\it set} of lexical items
is combined in a derivation step.  It is the elementary trees of the
underlying LTAG that determine the (extended local) domains over which
these dependencies can be expressed since it is the structure of an
elementary tree that determines the possible daughters in a
meta-production.  Although the types of elementary tree structures
licensed are specific to a particular LTAG, it might be expected that
a SLG(3) meta-grammar, for example, could encode the probability that
a given verb takes a particular (type of) subject and combination of
complements, including cases where the complements had been moved from
their canonical positions, for example by extraction. A meta-grammar
would also be likely to be able to differentiate the probabilities of
particular co-occurrences of adverbial and prepositional phrase
modifiers, and would moreover be able to distinguish between different
orderings of the modifiers.

The approach described by
\namecite{lafferty92} of associating probabilities with Link
Grammars---taken to its logical conclusion---corresponds to
SLG(3), since in that approach separate probabilities are associated with
each way of linking a word up with a combination of other
words\protect\footnote{\namecite{lafferty92} appear to
  consider only cases where a word has at most one right and one left
  link, i.e., probabilities are associated with at most
  triples. However, the formalism as defined by~\namecite{sleator93}
  allows a more general case with multiple links in each direction,
  as would be required to deal with, for example, modifiers.}.

\subsection{Globally-Dependent Frequencies}\label{sec-global}

The fourth and final approach we consider is Bod's Data-Oriented Parsing
(DOP) framework~\cite{bod95}. In this
paper we call it {\bf SLG(4)} for uniformity and ease of reference. Bod
formalizes DOP in terms of a {\it stochastic tree-substitution grammar},
which consists of a finite set of elementary trees, each with an associated
probability such that the probabilities of all the trees with the same
non-terminal symbol sum to $1$, with an operation of substitution to
combine the trees. In DOP, or SLG(4), the elementary trees are
arbitrarily large subtrees anchored at terminal nodes by words/part-of-speech
labels, and acquired automatically from pre-parsed training data. This
is in contrast to SLG(3), in which the size of individual meta-productions is
bounded, since the structure of the meta-productions is wholly determined
by the form of the elementary trees in the grammar.

\section{Empirical Evaluation}\label{sec-evaluation}

We have described four ways in which frequency information can be 
associated with a lexicalized grammar.  Directly comparing the performance
of the alternative schemes by training a wide-coverage grammar on an
appropriate annotated corpus and then parsing further, unseen data using
each scheme in turn would be a large undertaking outside the scope of this
paper.  However, each scheme varies in terms of the degree to which it can,
in principle, distinguish the probability of certain kinds of derivational
phenomena. This can be tested without the need to develop and run a parsing
system, since each scheme can be seen as making verifiable predictions about
the absence of certain dependencies in derivations of sentences in corpus
data.

SLG(1), with only context-free frequency information, predicts that
the relative frequency of use of the trees for a given nonterminal is
not sensitive to where the trees are used in a derivation.  For
example, there should be no significant difference between the
likelihood that a given NP tree is chosen for substitution at the
subject position and the likelihood that it is chosen for the object
position.  SLG(2) (using so-called node-dependent frequency
information) is able to cater for such differences but predicts that
the likelihood of substituting or adjoining a tree at a given node in
another tree is not dependent on what else is adjoined or substituted into
that tree.  With SLG(3) (which uses what we call locally-dependent
frequency information) it is possible to encode such sensitivity, but
more complex contextual dependencies cannot be expressed: for example, it is
not possible for the probability associated with the substitution or
adjunction of a tree $\gamma$ into another tree $\gamma'$ to be sensitive to
where the tree $\gamma'$ itself is adjoined or substituted. Only SLG(4)
(in which frequency information can be globally-dependent) can do this.

In the remainder of this section we present a number of empirical
phenomena that support or refute predictions made by each of the versions
of SLG.

\subsection{SLG(1) vs.\ SLG(2--4)}

\namecite{magerman91} report that, empirically, a noun
phrase is more likely to be realized as a pronoun in subject position than
elsewhere.  To capture this fact it is necessary to have two
different sets of probabilities associated with the different possible NP
trees: one for substitution in subject position, and another for
substitution in other positions.  This cannot be done in SLG(1) since
frequency information in SLG(1) is context-free. This phenomenon
therefore violates the predictions
of SLG(1), but it can be captured by the other SLG models.

Individual lexemes also exhibit these types of distributional
irregularities. For example, in the Wall Street Journal (WSJ) portion of the
Penn Treebank~2~\cite{Marcus94}, around 38\% of subjects of verbs used
intransitively (i.e., without an object NP) in active, ungapped
constructions are either pronouns or proper name phrases\footnote{Subjects
were identified as the NP-SBJ immediately preceding a VP bracketing
introduced by a verb labeled VBD/VBP/VBZ; pronouns, words labeled
PRP/PRP\$; and proper noun phrases, sequences of words all labeled
NNP/NNPS.}. However, for the verbs {\it believe}, {\it agree}, and {\it
understand}, there is a significantly higher proportion (in statistical
terms) of proper name/pronoun subjects (in the case of {\it believe} 57\%;
${\chi}^2$,~40.53, 1~$df$, $p<0.001$)~\footnote{A value for $p$ of 5
corresponds to statistical significance at the standard 95\%
confidence level; smaller values of $p$ indicate higher confidence.}.
This bias would, in semantic terms, be accounted for by a preference for
subject types that can be coerced to {\it human}. SLG(2--4) can
capture this distinction whereas SLG(1) cannot since it is not sensitive
to where a given tree is used.

\subsection{SLG(2) vs.\ SLG(3--4)}\label{sec-slg2}

The Penn Treebank can also allow us to probe the differences between the
predictions made by SLG(2) and SLG(3--4). From an analysis of verb
phrases in active, ungapped constructions with only pronominal and/or
proper name subjects and NP direct objects, it is the case that there is a
(statistically) highly significant dependency between the type of the
subject and the type of the object (${\chi}^2$,~29.79, 1~$df$, $p<0.001$),
the bias being towards the subject and direct object being either (a)~both
pronouns, or (b)~both proper names. Thus the choice of which type of NP
tree to fill subject position in a verbal tree can be dependent on the
choice of NP type for object position. Assuming that the subject and object
are substituted/adjoined into trees anchored by the verbs, this phenomenon
violates the predictions of SLG(2)---hence also SLG(1)---but can still
be modeled by SLG(3--4).

A similar sort of asymmetry occurs when considering the distribution of
pronoun and proper name phrases against other NP types in subject and
direct object positions. There is again a significant bias towards
the subject and object either both being a pronoun/proper name phrase, or
neither being of this type (${\chi}^2$,~8.77, 1~$df$, $p=0.3$). This
again violates the predictions of SLG(2), but not SLG(3--4).

Moving on now to modifiers, specifically prepositional phrase (PP) modifiers
in verb phrases, the Penn Treebank distinguishes several kinds including
PPs expressing manner (PP-MNR), time (PP-TMP), and purpose (PP-PRP). Where
these occur in combination there is a significant ordering effect: PP-MNR
modifiers tend to precede PP-TMP (${\chi}^2$,~4.12, 1~$df$, $p=4.2$), and
PP-TMP modifiers in their turn have a very strong tendency to precede
PP-PRP ($p<0.001$). Adopting Schabes and Shieber's~\shortcite{ss94}
formulation of the adjunction operation in TAG, multiple PP modifier trees
would be adjoined independently at the same node in the parent VP tree,
their surface order being reflected by their ordering in the derivation
tree. Therefore, in SLG(3) multiple modifying PPs would appear within a
single meta-production in the order in which they occurred, and the
particular ordering would be assigned an appropriate probability by virtue
of this. In contrast, SLG(2) treats multiple adjunctions separately and
so would not be able to model the ordering preference.

Significant effects involving multiple modification of particular
lexical items are also evident in the treebank. For example, the verb {\it
rise} occurs 83 times with a single PP-TMP
modifier---e.g.~\exnum{+1}{a}---and 12 times with two~\exnum{+1}{b},
accounting in total for 6\% of all PPs annotated in this way as
temporal.
\begin{ex}
\begin{subex}
{\it Payouts on the S\&P 500 stocks rose 10 \% [PP-TMP in 1988] , according
to Standard \& Poor 's Corp. ...}
\end{subex}
\begin{subex}
{\it It rose largely [PP-TMP throughout the session] [PP-TMP after posting
an intraday low of 2141.7 in the first 40 minutes of trading] .}
\end{subex}
\end{ex}
The proportion of instances of two PP-TMP modifiers with {\it rise} is
significantly more than would be expected given the total number of
instances occurring in the treebank (${\chi}^2$,~25.99, 1~$df$, $p<0.001$).
The verb {\it jump} follows the same pattern ($p=1.0$), but other synonyms
and antonyms of {\it rise} (e.g.\ {\it fall}) do not. This idiosyncratic
behavior of {\it rise} and {\it jump} cannot be captured by SLG(2), since
each adjunction is effectively considered to be a separate independent
event. In SLG(3), though, the two-adjunction case would appear in a single
meta-production associated with {\it rise/jump} and be accorded a higher
probability than similar meta-productions associated with other lexical
items.

There is another, more direct but somewhat less extensive, source of
evidence that we can use to investigate the differences between SLG(2)
and (3--4).  B.\ Srinivas at the University of Pennsylvania has recently
created a substantial parsed corpus\footnote{We wish to thank B.\ Srinivas
for giving us access to this resource.} by analyzing text from the Penn
Treebank using the XTAG system~\cite{xtag95}.  Some of the text has been
manually disambiguated, although we focus here on the most substantial
set---of some 9900 sentences from the WSJ portion---which has not been
disambiguated, as yet.  For each sentence we extracted the set of
meta-level productions that would generate the XTAG derivation.  To obtain
reliable data from ambiguous sentences, we retained only the (approximately
37500) productions that were common across all derivations.  In this set of
productions we have found that with the elementary tree licensing
subject--transitive-verb--object constructions, the likelihood that
the object NP is expanded with a tree anchored in {\it shares} is much
higher if the subject is expanded with with a tree anchored in {\it
volume}, corresponding to sentences such as~\exnum{+1}{a}
and~\exnum{+1}{b}.
\begin{ex}
\begin{subex}
{\it Volume totaled 14,890,000 shares .}
\end{subex}
\begin{subex}
{\it Overall Nasdaq volume was 151,197,400 shares .}
\end{subex}
\end{ex}
Indeed, in all 11
cases where {\it volume} is the anchor of the subject, an NP anchored in
{\it shares} is analyzed as the object, whereas more generally {\it shares}
is object in only 18 of the 1079 applications of the tree.  This difference
in proportions is statistically highly significant ($p<0.001$). 
Correlation between each of {\it volume} and {\it shares} and the verbs
that appear is much weaker.  There is of course potential for bias in the
frequencies since this data is based purely on unambiguous productions.  We
therefore computed the same proportions from productions derived from all
sentences in the XTAG WSJ data; this also resulted in a highly significant
difference. SLG(2) models the substitution of the subject and of the
object as two independent events, whereas the data show that they can
exhibit a strong inter-dependency.

\subsection{SLG(3) vs.\ SLG(4)}

\namecite{bod95} observes that there can be significant
inter-dependencies between two or more linguistic units, for example words
or phrases, that cut across the standard structural organization of a
grammar. For example, in the Air Travel Information System
(ATIS) corpus~\cite{hemphill90} the generic noun phrase (NP) {\it
flights from X to Y} (as in sentences like {\it Show me flights from Dallas to
Atlanta}) occurs very frequently. In this domain the dependencies between
the words in the NP---but without {\it X} and {\it Y} filled in---are so
strong that in ambiguity resolution it should arguably form a single
statistical unit. Bod argues that Resnik and Schabes' schemes (i.e.\
SLG(2)) cannot model this; however it appears that SLG(3) can since the NP
would give rise to a single meta-production (under the reasonable assumption
that the {\it from} and {\it to} PPs would be adjoined into the NP
tree anchored by {\it flights}).

An example given by Bod that does demonstrate the difference between SLG(3)
and SLG(4) concerns sentences like {\it the emaciated man starved}. Bod
argues that there is a strong (semantic) dependence between {\it emaciated}
and {\it starved}, which would be captured in DOP---or SLG(4)---in the form
of a single elementary tree in which {\it emaciated} and {\it starved} were
the only lexical items. This dependence cannot be captured by SLG(3)
since {\it emaciated} and {\it starved} would anchor separate elementary
trees, and the associations made would merely be between (1)~the S
tree anchored by {\it starved} and the substitution of the NP anchored by
{\it man} in subject position, and (2)~the modification of {\it man} by {\it
emaciated}.

\subsection{Discussion}

The empirical phenomena discussed above mainly concern interdependencies
within specific constructions between the types or heads of either
complements or modifiers. The phenomena fall clearly into two
groups:
\begin{itemize}
\item ones relating to distributional biases that are independent of
particular lexical items, and
\item others that are associated with specific open class vocabulary.
\end{itemize}
Token frequencies---with respect to treebank data---of phenomena in the
former group are relatively high, partly because they are not
keyed off the presence of a particular lexical item: for example in the
case study into the complement distributions of pronoun/proper name phrases
versus other NP types (section~\ref{sec-slg2}) there are
13800 data items (averaging one for every four treebank sentences).
However, there appears to be a tendency for the phenomena in this group to
exhibit smaller statistical biases than are evident in the latter,
lexically-dependent group (although all biases reported here are
significant at least to the 95\% confidence level). In the latter group,
although token frequencies for each lexical item are not large (for
example, the forms of {\it rise} under consideration make up only 1\% of
comparable verbs in the treebank), the biases are in general very strong,
in what are otherwise an unremarkable set of verbs and nouns ({\it
believe}, {\it agree}, {\it understand}, {\it rise}, {\it jump}, {\it
volume}, and {\it shares}). We might therefore infer that although
individually token frequencies are not great, {\it type} frequencies are
(i.e.\ there are a large number of lexical items that display idiosyncratic
behavior of some form or other), and so lexicalized interdependencies are
as widespread as non-lexical ones.

\section{Parameter Estimation}\label{sec-estimation}

\subsection{Training Regime}

\namecite{schabes:1992} describes an iterative re-estimation
procedure (based on the Inside-Ouside Algorithm~\cite{baker79})
for refining the parameters of an SLG(2) grammar given a corpus of
in-coverage sentences; the algorithm is also able to simultaneously acquire
the grammar itself. The aim of the algorithm is to distribute the
probability mass within the grammar in such as way that the
{\it probability of the training corpus} is maximized, i.e.\ model as closely
as possible the language in that corpus. However, when the goal is to return
as accurately as possible the {\it correct analysis} for each sentence using
a pre-existing grammar, estimating grammar
probabilities directly from normalized frequency counts derived from a
pre-parsed training corpus can result in accuracy that is comparable or
better to that obtained using re-estimation~\cite{charniak96}. Direct
estimation would mesh well with the SLG formalisms described in
this paper.

\subsection{Smoothing}
\label{sec-smoothing}

The huge number of parameters required for a wide-coverage SLG(2) (and
even more so for SLG(3--4)) means that not only would the amount
of frequency information be unmanageable, but data sparseness would make
useful probabilities hard to obtain. We briefly present three (essentially
orthogonal and independent) backing-off techniques that could be used to
address this problem.

\subsubsection{Unanchored Trees}

It is the size of a wide-coverage lexicon that makes pure SLG(2--4)
unmanageable. However, without lexical anchors a wide-coverage SLG would
have only a few hundred trees \cite{xtag95}. Backup frequency values could
therefore be associated with unanchored trees and used when data for the
anchored case was absent.

\subsubsection{Lexical Rules}

In a lexicalized grammar, elementary trees may be grouped into families
which are related by lexical rules---such as {\it wh} extraction, and
passivization. (For example, the XTAG grammar contains of the order of 500
rules grouped into around 20 families). In the absence of specific frequency
values, approximate (backup) values could be obtained from a tree that
was related by some lexical rule.

\subsubsection{SLG($i$) to SLG($i-1$)}

Section~\ref{sec-evaluation} indicated informally how, when moving from
SLG(1) through to SLG(4), the statistical model becomes successively more
fine-grained, with each SLG($i$) model subsuming the previous ones, in the
sense that SLG($i$) is able to differentiate probabilistically all
structures that previous ones can. Thus, when there is insufficient
training data, sub-parts of a finer-grained SLG model could be backed off
to a model that is less detailed. For example, within a SLG(3)
model, in cases where a particular set of meta-productions all with the same
mother had a low collective probability, the set could be reduced to a single
meta-production with unspecified daughters (i.e.\ giving the effect of
SLG(1)).

\section{Comparison with Other Work}\label{sec-other}

The treatment of stochastic lexicalized grammar in this paper has much in
common with recent approaches to statistical language modeling outside the
TAG tradition. Firstly, SLG integrates statistical preferences acquired from
training data with an underlying wide-coverage grammar, following an
established line of research, for
example~\cite{chitrao90,charniak94,briscoe95}. The paper discusses
techniques for making preferences sensitive to context to avoid known
shortcomings of the context-independent probabilities of SCFG (see e.g.\
\namecite{briscoe93}).

Secondly, SLG is {\it lexical}, since
elementary trees specify lexical anchors.
Considering the anchor of each elementary
tree as the head of the construction analyzed, successive daughters for
example of a single SLG(3) meta-grammar production can in many cases
correspond to a combination of Magerman's~\shortcite{magerman95}
mother/daughter and daughter/daughter head statistics (although it would
appear that Collins'~\shortcite{collins96} head-modifier configuration
statistics are equivalent only to SLG(2) in power).  However, due to its
extended domain of locality, SLG(3) is not limited to modeling local
dependencies such as these, and it can express dependencies between heads
separated by other, intervening material.  For example, it can deal directly
and naturally with dependencies between subject and any verbal complement
without requiring mediation via the verb itself: c.f.\ the example of
section~\ref{sec-slg2}.

Thirdly, the SLG family has the ability to model
explicitly syntactic
{\it structural} phenomena, in the sense that the atomic structures to which
statistical measures are attached can span multiple levels of derived parse
tree structure, thus relating constituents that are
widely-separated---structurally as well as sequentially---in a sentence.
Bod's DOP model~\cite{bod95} shares this characteristic, and indeed (as
discussed in section~\ref{sec-global}) it fits naturally into this family,
as what we have called SLG(4).

\namecite{srinivas96} (see also \namecite{joshi94}) have recently described a
novel approach to parsing with LTAG, in which each word in a sentence is
first assigned the most probable elementary tree---or `supertag'---given the
context in which the word appears, according to an trigram model of
supertags. The rest of the parsing process then reduces to finding a way of
combining the supertags to form a complete analysis. In this approach
statistical information is associated simply with linear sub-sequences of
elementary trees, rather than with trees within derivational contexts as in
SLG(2--4). Although Srinivas' approach is in principle efficient,
mistaggings mean that it is not guaranteed to return an analysis for every
in-coverage sentence, in contrast to SLG. Also, its relatively impoverished
probabilistic model would not be able to capture many of the phenomena
reported in section~\ref{sec-evaluation}.

\section*{Acknowledgement}

This work was supported by UK EPSRC project GR/K97400 `Analysis of
Naturally-occurring English Text with Stochastic Lexicalized Grammars'
({\tt <http://www.cogs.susx.ac.uk/lab/nlp/dtg/details.html>}), and by an
EPSRC Advanced Fellowship to the first author. We would like to thank Nicolas
Nicolov and Miles Osborne for useful comments on previous drafts.


\end{document}